\pdfoutput=1

\documentclass[11pt]{article}

\usepackage[final]{acl}

\usepackage{times}
\usepackage{latexsym}

\usepackage[T1]{fontenc}

\usepackage[utf8]{inputenc}

\usepackage{microtype}

\usepackage{inconsolata}

\usepackage{graphicx}
\usepackage{algorithm}
\usepackage{algorithmic}
\usepackage{array} 
\usepackage{booktabs}

\usepackage{amsmath}
\usepackage{appendix}
\usepackage{xcolor}
\usepackage{CJKutf8}
\usepackage{multirow}
\usepackage{amssymb}
\usepackage{nameref}
\usepackage{hyperref}
\usepackage{caption}
\usepackage{tcolorbox}

\newcommand{\ourmethod}{SynapticRAG}

\title{SynapticRAG: Enhancing Temporal Memory Retrieval \\in Large Language Models through Synaptic Mechanisms}

\author{
    Yuki Hou\textsuperscript{1}, 
    Haruki Tamoto\textsuperscript{2}, 
    Qinghua Zhao\textsuperscript{3}, 
    Homei Miyashita\textsuperscript{1}\\
    \textsuperscript{1}Meiji University, 
    \textsuperscript{2}Kyoto University, 
    \textsuperscript{3} Hefei University\\
    \texttt{houhoutime@gmail.com, harukiririwiru@gmail.com, zhaoqh@buaa.edu.cn}
}

\begin{document}
\maketitle
\begin{abstract}
Existing retrieval methods in Large Language Models show degradation in accuracy when handling temporally distributed conversations, primarily due to their reliance on simple similarity-based retrieval. Unlike existing memory retrieval methods that rely solely on semantic similarity, we propose \ourmethod, which uniquely combines temporal association triggers with biologically-inspired synaptic propagation mechanisms. Our approach uses temporal association triggers and synaptic-like stimulus propagation to identify relevant dialogue histories. A dynamic leaky integrate-and-fire mechanism then selects the most contextually appropriate memories. Experiments on four datasets of English, Chinese and Japanese show that compared to state-of-the-art memory retrieval methods, \ourmethod~achieves consistent improvements across multiple metrics up to $14.66\%$ points. This work bridges the gap between cognitive science and language model development, providing a new framework for memory management in conversational systems. \footnote{The code is available at: \url{https://github.com/tamoharu/SynapticRAG-Benchmark}}
\end{abstract}


\section{Introduction}
As AI systems become increasingly integrated into daily communications, the ability to maintain natural conversations has become crucial \citep{Dai2019TransformerXLAL,trivedi-etal-2023-interleaving,deng2024}. While Large Language Models (LLMs) have achieved remarkable progress in dialogue generation, they face a fundamental challenge: maintaining coherent long-term memory \citep{NEURIPS2020_c8512d14,NEURIPS2020_1457c0d6,beltagy2020longformerlongdocumenttransformer}. This limitation becomes particularly evident in their struggle to retrieve and integrate contextually relevant information from temporally distributed dialogue histories \citep{endtoend,gao-etal-2018-neural-approaches,olabiyi-etal-2019-multi,hipporag}. For instance, current state-of-the-art memory retrieval methods show a 17\% degradation in response relevance when conversation context spans multiple sessions, significantly impacting user experience in real-world applications.

Inspired by cognitive science research, which shows that human memory retrieval relies heavily on temporal triggers \citep{Tul1985,Mcdanie} and synaptic connections between memory nodes \citep{hebb,OReilly1994HippocampalCE,Norman2003,neves2008synaptic}, rather than pure memory strength. In contrast, current methods primarily use simple similarity-based retrieval, failing to capture these temporal-associative aspects.

Inspired by these cognitive principles, we propose \ourmethod, a neuroscience-inspired memory retrieval framework. Our approach integrates temporal association triggers, synaptic-like memory connections, and dynamic activation control to achieve more human-like memory retrieval in dialogue systems.  This approach is theoretically motivated by the proven effectiveness of biological memory systems in handling temporal dependencies and context integration. Our work makes three key contributions: (1) A novel biologically-inspired memory retrieval mechanism, (2) Integration of temporal and associative memory aspects in memory retrieval systems, and (3) Empirical validation across three languages and datasets, showing consistent improvements up to $14.66\%$ points over state-of-the-art memory retrieval methods on English, Chinese and Japanese benchmarks.

\vspace{-0.1cm}

\section{Related Works}
\vspace{-0.1cm}
\subsection{Memory Retrieval}
Traditional retrieval-augmented generation (RAG) approaches focus primarily on semantic similarity for knowledge retrieval \citep{rag,rag2,rag3}. While effective for static knowledge bases, these methods often struggle with maintaining dialogue coherence over extended interactions. Recent studies have explored various temporal enhancement strategies: \citet{memorybank} and \citet{alonso2024conversationalagentscontexttime} introduced time-based memory updates, while \citet{myagent} demonstrated the benefits of dynamic memory consolidation. Despite these advances, significant challenges remain in temporal aspects of task-oriented dialogues \citep{wu2019learning,zhang2018memory}. The latest developments in self-updatable models \citep{liu2024self} and behavioral memory \citep{lee-etal-2024-improving-conversational} underscore the importance of integrating semantic and temporal information, which our work directly addresses.

\subsection{Biological Inspiration}
Neuroscience research has revealed that synaptic mechanisms play crucial roles in memory formation and retrieval \citep{Gerstner2002SPIKINGNM}. Particularly relevant to our work is the phenomenon of synaptic reverberation in working memory maintenance \citep{Compte2000,Wang2001}. While existing dialogue systems often employ simplified decay mechanisms, biological studies suggest the importance of diverse temporal dynamics in memory processes \citep{fell2011,Howard2024}. Our \ourmethod~draws upon these insights by implementing biologically-motivated synaptic propagation for temporal-aware memory retrieval, addressing the limitations of conventional approaches that lack biological grounding.

\begin{figure*}[t]
\small
\centering
\includegraphics[width=\textwidth]
{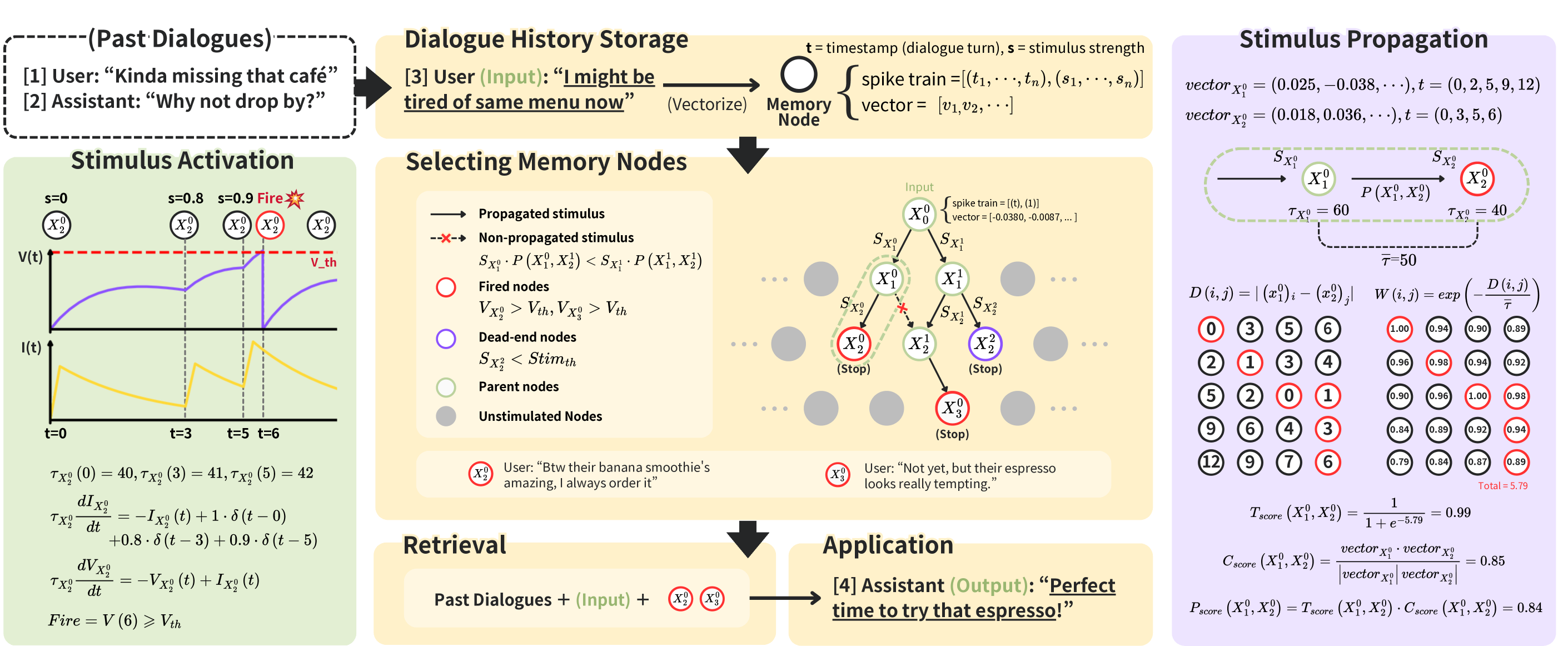}
\caption{The framework of SynapticRAG. First, past dialogues and current input are stored as memory nodes and formalized into an attributed graph; Second, the stimulus strength is propagated among the graph following propagation mechanisms; Third, nodes are activated through temporal-semantic connections following neuron-inspired dynamics; Finally, responses are generated by integrating selected memories with the input query.}
\label{fig:synapticRAG}
\end{figure*}

\section{Approach}
Our approach enhances dialogue memory recall by integrating temporal dynamics into retrieval mechanisms. The model consists of four components, as described in Figure \ref{fig:synapticRAG}.

\subsection{Memory Construction and Storage}
We embed each dialogue turn into a vector representation and store them as memory nodes in a graph structure. The edges between nodes are weighted by their cosine similarities, reflecting the semantic relationships in the dialogue history. Each node maintains a weighted spike train $[(t_i, s_i)]_{i=1}^n$, where $t_i$ represents the dialogue turn number, $s_i$ denotes the stimulus strength (initialized to 1), and $n$ counts the received stimuli. This temporal information enables our model to capture both the semantic relevance and temporal patterns of dialogue interactions.

\subsection{Selecting Memory Nodes}
Our selection process combines semantic and temporal filtering to identify relevant memory nodes. Given a new query, we first select candidate nodes whose semantic similarity exceeds threshold $\text{cos}_\text{th}$. These candidates are then filtered based on temporal association strength, leveraging the observation that temporally proximate memories often form coherent dialogue chains \citep{tempmemorycite1,tempmemorycite2}.

\paragraph{Temporal Association Computation}
We quantify temporal associations between memory nodes using a framework based on Dynamic Time Warping (DTW) \cite{dtwJP}, which enables robust alignment of memory spike trains across time and supports flexible modeling of non-linear temporal shifts. For two memory nodes with spike trains $A = (t^a_1, ..., t^a_n)$ and $B = (t^b_1, ..., t^b_m)$, where $m, n$  denotes the largest turn number, we construct a distance matrix $D \in \mathbb{R}^{n \times m}$ where each element $D(i, j) = |t^a_i - t^b_j|, 1\leq i\leq n, 1\leq j\leq m$ represents the absolute temporal difference between stimulus reception timestamps.

To transform temporal distances into association strengths, we apply an exponential decay using the average time constant $\bar{\tau} = 0.5(\tau_A + \tau_B)$, where $\tau_A$ and $\tau_B$ are node-specific time constants, we construct the association strength matrix $W \in \mathbb{R}^{n \times m}$ as:
\begin{equation}
W(i, j) = \exp\left(\frac{-D(i, j)}{\bar{\tau}}\right) \in (0,1], 
\end{equation}

where larger $\bar{\tau}$ values maintain robust memory connections with sustained information retention, while smaller values lead to rapid decay. The exponential transformation bounds association strengths within $(0,1]$, with stronger temporal proximity reflected in higher scores.

Given $W$, we maximize the cumulative association strength by defining cumulative matrix $L$ as:
\vspace{-0.7cm}
\begin{align}
L(i,1) &= \sum_{k=1}^i W(k,1), \quad i \in \{1,...,n\} \\
L(1,j) &= \sum_{k=1}^j W(1,k), \quad j \in \{1,...,m\}
\end{align}
\begin{equation}
\begin{split}
L(i,j) &= W(i,j) + \max\{L(i-1,j),\,\\
       &\quad L(i,j-1),\,L(i-1,j-1)\}
\end{split}
\end{equation}

The cumulative score $L(n,m)$ uniquely captures the temporal association strength between sequences $A$ and $B$. 

Besides, as $L(n,m)$ may exceed $1$, we normalize this cumulative score through sigmoid transformation $\sigma(x)=\frac{1}{1 + \exp(-x)}$:

\begin{equation}
T_{\text{score}}(A,B) = \sigma(L(n,m))
\end{equation}

\paragraph{Score Integration} $T_{\text{score}}$ captures the temporal association strength between memory nodes. To integrate temporal and semantic similarities into a unified association measure, we define the propagation score $P_{\text{score}}$ as:

\begin{equation}
\label{pscore}
P_{\text{score}}(A,B) = T_{\text{score}}(A,B) \cdot C_{\text{score}}(A,B)
\end{equation}

where $C_{\text{score}}(A,B) = \frac{A \cdot B}{|A| \cdot |B|}$ quantifies semantic similarity through cosine distance between node vectors. This integration provides a comprehensive measure for subsequent stimulus propagation.

\subsection{Stimulus Propagation Mechanism}\label{spm}
We organize memory nodes into a layered graph for efficient stimulus propagation. In layer $i$, nodes that accumulate sufficient stimulus become active and can propagate to connected nodes in layer $i+1$. The propagation strength between nodes follows $P_{\text{score}}$ (Eq. \eqref{pscore}), forming contextually coherent memory paths. The propagation strength between parent node $X_i$ and its child node $X_{i+1}$ follows:

\begin{equation}
S_{X_{i+1}} = P_{\text{score}}(X_i, X_{i+1}) \cdot S_{X_i}, \quad S_{X_0} = 1
\end{equation}

where $S_{X_{i+1}}$ denotes the stimulus strength received at node $X_{i+1}$, $P_{\text{score}}(X_i, X_{i+1})$ quantifies the temporal and semantic association strengths between parent nodes, $S_{X_i}$ denotes the current stimulus strength at node $X_i$, and $S_{X_0}$ establishes the baseline stimulus strength for the initial input node.

Memory nodes receive activation signals that propagate along paths in the layered graph. For a propagation path $X_0, X_1, ..., X_n$, the stimulus strength at node $X_n$ decays according to:
\begin{equation}
\label{eq8}
S_{X_n} = S_{X_0} \prod_{i=0}^{n-1} P_{\text{score}}(X_i, X_{i+1})
\end{equation}
where $S_{X_0}$ is the initial stimulus strength and $P_{\text{score}}$ captures the decay along each step.

\paragraph{Stimulus Propagation Path Construction}
We construct propagation paths through three steps: parent node qualification, child node selection, and path formation. Let $\mathcal{N}$ denote the universal node set, with thresholds $\text{stim}_\text{th}$ and $\text{cos}_\text{th}$ controlling activation and connection sensitivity respectively.

The propagation process follows a layer-wise approach, where qualified parent nodes $\mathcal{P}_i$ at layer $i$ are determined by:

\begin{equation}
\mathcal{P}_i = \{p \in \mathcal{N}_i \mid (S_p > \text{stim}_\text{th}) \land (p \notin \text{Fire})\}
\end{equation}

where $S_p$ denotes the stimulus strength. The thresholds are optimized as detailed in Appendix \ref{hyperparameterOpt}, with Fire Mechanism described in Section \ref{DLIF}.

For each parent node $p\in\mathcal{P}_i$, we define its candidate child node set $\mathcal{C}_i({p})$ for layer $i+1$. $\mathcal{C}_i({p})$ is constructed as the set of nodes whose cosine similarity with parent node $p$ exceeds threshold $\text{cos}_\text{th}$, while excluding all nodes from layers 0 through $i$:

\begin{equation}
\mathcal{C}_i(p) = \left\{c\in \mathcal{N} \setminus \bigcup_{j=0}^i \mathcal{N}_j \mid \frac{p \cdot c}{\|p\| \|c\|} > \text{cos}_\text{th} \right\}
\end{equation}

Based on the candidate child node set $\mathcal{C}_i(p)$, we determine the actual stimulus-receiving child nodes. For cases where a child node $c$ receives stimuli from multiple parent nodes, we select the parent node providing the maximum stimulus strength and utilize its stimulus value.

We first formalize a candidate parent node set $\mathcal{P}_i(c)$ for each child node as:
\begin{equation}
\mathcal{P}_i(c) = \left\{ p \in \mathcal{P}_i \mid \frac{p \cdot c}{\|p\| \|c\|} > \text{cos}_\text{th} \right\}
\end{equation}

The optimal parent node $p^*$ that transmits stimulus to child node $c$ is uniquely determined from $\mathcal{P}_i(c)$ through:
\begin{equation}\label{eq:parent-node}
p^* = \arg\max_{p \in \mathcal{P}_i(c)} S_p \cdot P_{\text{score}}(p,c)
\end{equation}

We formalize the actual child node set $\mathcal{C}_i^*$ as:
\begin{equation}
\begin{split}
\mathcal{C}_i^*(p) = \Bigl\{ c \in \mathcal{C}_i(p) \mid \, & S_p \cdot P_{\text{score}}(c, p) \\
& = \max_{p' \in \mathcal{P}_i} S_p \cdot P_{\text{score}}(p',c) \Bigr\}
\end{split}
\end{equation}

The node set $\mathcal{N}_{i+1}$ at layer $i+1$ is constructed as the union of all actual child node sets $\mathcal{C}_i^*(p)$, with the constraint that these child node sets must be mutually disjoint for different parent nodes:
\begin{equation}
\begin{split}
\mathcal{N}_{i+1} = \bigcup_{p \in \mathcal{P}_i} \mathcal{C}_i^*(p),
\mathcal{C}_i^*(p) \cap \mathcal{C}_i^*(q) &= \emptyset\\ \quad (\forall p, q \in \mathcal{P}_i, \, p \neq q)
\end{split}
\end{equation}

With $\mathcal{N}_{i+1}$ established for layer $i+1$, we iteratively execute this process for subsequent layers. The propagation terminates when there are no further qualified parent nodes, specifically when $\mathcal{P}_i = \emptyset$. Following this propagation framework, we can define the pathway to node $X_n$ using the method for determining parent nodes mentioned earlier, as expressed by the following equation:
\begin{equation}
X_i = \arg\max_{x \in \mathcal{P}_i(X_{i+1})} S_p \cdot P_{\text{score}}(x,X_{i+1})
\end{equation}

\subsection{Stimulus Activation}
\label{DLIF}
Based on the stimulus propagation mechanism (Section \ref{spm}), we formalize how nodes process received stimuli through membrane potential: a dynamic variable that accumulates input stimuli and governs node activation. For a node $X$, the processing pipeline consists of three steps: (1) converting discrete stimuli into continuous input current, (2) evolving membrane potential with dynamic time constants, and (3) determining activation through threshold-based firing. This framework enables adaptive memory selection in dialogue systems by emulating neurobiological principles \citep{lif2}.

\paragraph{Input Current Generation}
The Leaky Integrate-and-Fire (LIF) model \cite{lif1} provides a biologically inspired abstraction for simulating how stimulus input accumulates and decays over time within memory nodes, forms the basis for processing temporal signals in dialogue systems, however, its fixed membrane time constant $\tau$ limits its ability to balance recent inputs with long-term memories in dialogue systems. To alleviate this, we propose an enhanced LIF model with a dynamic time constant $\tau_X$ and a spike train-based input accumulation mechanism. The discrete input stimuli $S_X(t)$ are converted into continuous input current $I_X(t)$ through:

\begin{equation}\label{eq:input-current}
\tau_X \frac{dI_X}{dt} = -I_X(t) + \sum_{t_s \in \Gamma} S_X(t) \delta(t - t_s)
\end{equation}

where $\Gamma$ denotes input pulse timestamps, $t_s$ represents stimulus reception, and $\delta(t - t_s)$ is the Dirac delta function for discrete pulse modeling \citep{bracewell2000fourier}. The leakage term $-I_X(t)$ simulates neuronal ion channels, enabling natural decay of past stimuli while maintaining responsiveness to new inputs. When no stimulus is present, the summation term becomes zero, allowing gradual memory decay analogous to biological systems.

\paragraph{Membrane Potential Evolution}
Node state is represented by membrane potential $V_X(t)$,  reflecting both input stimulus accumulation and natural memory decay, which evolves according to:

\begin{equation}
\tau_X \frac{dV_X}{dt} = -V_X(t) + I_X(t)
\end{equation}

where the decay term $-V_X(t)$ prevents unbounded accumulation while $I_X(t)$ incorporates new inputs. This evolution enables selective activation as nodes fire only when $V_X(t)$ exceeds a predefined threshold. To balance memory retention with input sensitivity, we introduce a dynamic membrane time constant $\tau_X$ that updates based on input intervals $\Delta t$:

\begin{equation}\label{eq:tau}
\tau_X(t + \Delta t) = \tau_X(t) + \frac{1 - \exp(-\Delta t)}{1 + \exp(-\Delta t)}
\end{equation}

This sigmoid mapping ensures that $\tau_X$ remains stable (approaches 0) for frequent inputs to maintain sensitivity, while increasing (approaches 1) for sparse inputs to preserve long-term memories. $\tau_X$ starts with its optimaized initial value $\tau_{\text{init}}$, which is fixed for all nodes. This mechanism effectively emulates synaptic plasticity in biological neurons.

\paragraph{Node Activation and Fire Mechanism}
A node enters the Fired state when its membrane potential exceeds threshold $V_{\text{th}}$, with the firing nodes at time $t$ defined as $\text{Fire}(t) = \{X \in \mathcal{N}\mid V_X(t) \geq V_{\text{th}}\}$.
Post-firing, nodes reset to optimized resting states $(V_{\text{rest}}, I_{\text{rest}})$ (See Appendix \ref{appendixB}), preventing continuous activation and enabling contextually relevant memory selection. The dialogue text of nodes in fired state is incorporated into the prompt for response generation, thus contributing to the system's contextual memory retrieval.

\section{Experiments}
\subsection{Dataset} 
Temporal information is critical for dialogue response generation, requiring access to historical context for time-dependent interactions. Effective evaluation of temporal-aware dialogue capabilities requires datasets with three key properties: (1) multi-turn dialogues that encompass complex temporal contexts spanning multiple interactions across distinct temporal segments, (2) complex temporal dependencies where correct memory retrieval depends critically on temporal context rather than solely semantic similarity, and (3) explicit annotations identifying correct memory retrievals, providing ground-truth labels for quantitatively evaluating retrieval accuracy. Currently, only \textbf{PerLTQA} \citep{du-etal-2024-perltqa} partially addresses these requirements through its format with temporal dependencies, containing 625 English and Chinese dialogues.

To fill this gap, we introduce Synaptic Memory Retrieval Conversations \textbf{SMRCs}, a bilingual dataset in English and Japanese consisting of 101 dialogues with 456 tasks, constructed using Claude 3.5 Sonnet \cite{claude3.5} (See generation guidelines in Appendix \ref{guidelines}). Each task consists of two segments: a past and a present dialogue, with 10 turns of user-assistant interactions each.

For quantitative evaluation, we annotate each task with multiple trigger-memory pairs, where a trigger is an utterance in the present dialogue referencing the past context, and memories are relevant utterances from the past dialogue. To maintain natural conversational flow and consistent information density, utterances are controlled at lengths between 10 and 20 words. Temporal relationships between triggers and memories typically span multiple turns, requiring models to comprehend dialogue contexts beyond superficial semantic matching. Figure \ref{fig:case} provides a representative example.

\subsubsection{Annotation Reliability Evaluation}
To ensure semantic authenticit and annotation validity, we implemented a rigorous filtering method for dataset construction. Specifically, annotations were independently performed by four annotators, and only annotations meeting one of two strict conditions were retained: (1) annotations were exactly identical, with at least two memory references; or (2) annotations partially overlapped, and the intersection contained at least two memory references, retaining only the overlapping annotations.

\begin{table}[h]
\small
\centering
\caption{Annotation results from four annotators (A1–A4) across 10 tasks. Each cell indicates the indices of selected utterances.}
\label{tab:annotation_results}
\begin{tabular}{ccccc}
\toprule
\textbf{Task ID} & \textbf{A1} & \textbf{A2} & \textbf{A3} & \textbf{A4} \\
\midrule
1 & [4,5] & [4,5] & [5,6] & [4,5] \\
2 & [6,7] & [6,7] & [4,6] & [6,7] \\
3 & [2,4] & [2,4] & [2,4] & [2,4] \\
4 & [8,9] & [8,9] & [8,9] & [6,8] \\
5 & [0,2] & [0,2] & [0,2] & [0,2] \\
6 & [5,7] & [5,7] & [5,7] & [5,7] \\
7 & [2,7] & [2,7] & [2,7] & [2,6] \\
8 & [6,8] & [6,8] & [8,9] & [8,9] \\
9 & [5,6] & [5,6] & [5,6] & [5,6] \\
10 & [0,7] & [0,7] & [0,6] & [6,7] \\
\bottomrule
\end{tabular}
\end{table}

Furthermore, we conducted an inter-annotator agreement evaluation using Krippendorff's Alpha \cite{krippendorff2011computing}. We selected 10 representative tasks from the dataset, covering diverse dialogue scenarios. Annotators independently followed clearly defined annotation guidelines: each annotator was given a set of 10 past dialogue utterances and one trigger utterance from the present dialogue and instructed to select 2 past utterances relevant for accurately comprehending and responding to the trigger. For calculating agreement, we constructed a binary selection matrix representing each of the 10 past utterances per task, assigning a value of 1 if selected by an annotator and 0 otherwise. This matrix was analyzed using a nominal scale to compute inter-annotator agreement. The annotation results are summarized in Table \ref{tab:annotation_results}.

Based on these annotations, we computed a Krippendorff's Alpha agreement value of 0.782. Typically, a Krippendorff's Alpha value $\geq$ 0.8 indicates excellent reliability, while a value $\geq$ 0.67 is considered an acceptable minimum standard \cite{krippendorff04}. This result demonstrates robust consistency and strong consensus among annotators, with limited subjectivity in the annotations.

\begin{figure*}[t]
\centering
\small
\includegraphics[width=\textwidth]{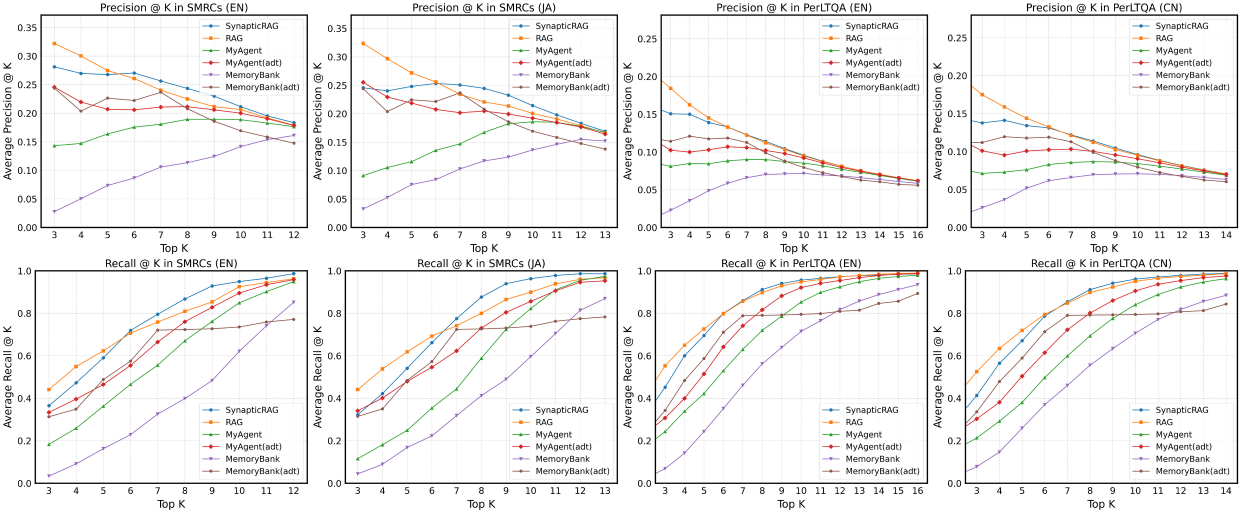}
\caption{Traditional memory retrieval metrics values (Precision@$k$ and Recall@$k$, $k=[3,\ldots,12]$) across four tested datasets and five baselines.}
\label{fig:k_values}
\end{figure*}

\subsection{Baselines and Model Adaptations}
We compare \ourmethod~with five baseline models. The three base models are standard RAG, MemoryBank \cite{memorybank}, MyAgent \cite{myagent}, and our SynapticRAG. For fair comparison against SynapticRAG's temporal-aware retrieval mechanism, we also developed adapted versions of MemoryBank and MyAgent.

\textbf{RAG} serves as our fundamental baseline, as its vector similarity search forms the foundation of retrieval process and provides a robust benchmark for evaluating basic retrieval capabilities. 

\textbf{MemoryBank} incorporates Ebbinghaus Forgetting Curve-inspired \cite{graves2016hybrid} mechanisms to dynamically adjust memory based on temporal factors, representing an improved memory management approach. 

\textbf{MyAgent} represents a cognitive-inspired approach to memory retrieval, adopting human memory cue recall as a trigger and developing a mathematical model that dynamically quantifies memory consolidation by considering contextual relevance $r$, elapsed time $t$, and recall probability $p$.

\textbf{MemoryBank (Adt)} enhances the original MemoryBank's temporal awareness through additional parameters including top-$K$, $\Theta$, and scaling factors ($t_{\text{scale}}$, $s_{\text{scale}}$, $s_{\text{init}}$) to enable fair comparison with temporal-aware retrieval mechanisms. 

\textbf{MyAgent (Adt)} modifies the original retrieval process to select the same number of memories as \ourmethod~from those exceeding the cosine similarity threshold, ranked by recall probability $p$, ensuring consistent comparison conditions.

\begin{table*}[ht]
\centering
\caption{Dynamic memory retrieval metrics values (ERC and ERC-MG). $\Delta$ denotes the improvement of \ourmethod~over the best baseline model.}
\small
\begin{tabular}{lccccccccc}
\toprule
\multirow{2}{*}{} & \multicolumn{2}{c}{SMRCs\textsuperscript{EN}} & \multicolumn{2}{c}{SMRCs\textsuperscript{JA}} & \multicolumn{2}{c}{PerLTQA\textsuperscript{EN}} & \multicolumn{2}{c}{PerLTQA\textsuperscript{CN}} & \multirow{2}{*}{Avg.} \\
\cmidrule(lr){2-3} \cmidrule(lr){4-5} \cmidrule(lr){6-7} \cmidrule(lr){8-9}
& ERC & ERC-MG & ERC & ERC-MG & ERC & ERC-MG & ERC & ERC-MG & \\
\midrule
RAG & 71.55 & 71.55 & 81.58 & 81.58 & 82.14 & 82.51 & 84.95 & 85.36 & 80.15 \\
MemoryBank & 32.60 & 32.82 & 49.34 & 49.34 & 33.62 & 33.62 & 36.75 & 36.75 & 38.11 \\
MemoryBank\textsubscript{(Adt)} & 62.36 & 62.58 & 72.15 & 72.15 & 67.24 & 67.54 & 71.31 & 71.61 & 68.37 \\
MyAgent & 62.14 & 62.14 & 70.61 & 70.61 & 61.99 & 62.29 & 65.14 & 65.47 & 65.05 \\
MyAgent\textsubscript{(Adt)} & 63.24 & 63.46 & 77.19 & 77.19 & 66.49 & 66.64 & 69.54 & 69.76 & 69.19 \\
\midrule
\textbf{SynapticRAG} & \textbf{86.21} & \textbf{86.21} & \textbf{92.32} & \textbf{92.32} & \textbf{90.47} & \textbf{90.47} & \textbf{93.12} & \textbf{93.12} & \textbf{90.53} \\
$\Delta$\% & +14.66 & +14.66 & +10.74 & +10.74 & +8.33 & +7.96 & +8.17 & +7.76 & +10.38 \\
\bottomrule
\end{tabular}
\label{tab:retrieval_performance}
\end{table*}

\subsection{Metrics}
We evaluate \ourmethod~using both traditional and dynamic retrieval metrics. Traditional metrics like recall@$k$ and precision@$k$ measure retrieval performance with a fixed $k$ number of memories \citep{Manning_Raghavan_Schütze_2008,10184013,10.5555/1516224}. Unlike the baseline methods, SynapticRAG does not inherently retrieve a fixed number of memories. To fairly compute and compare Precision and Recall metrics, we implemented the following procedure:

For each retrieval scenario, we ranked memory nodes according to their membrane potentials ($V$) computed by SynapticRAG. Subsequently, we selected the top-$k$ memory nodes with the highest $V$ values to match the number of memories retrieved by the baseline methods. Precision and Recall were then calculated based on these top-$k$ selected memories. Since $V$ in SynapticRAG directly determines whether a node fires (i.e., is retrieved), ranking memories by membrane potential reflects the model’s intrinsic memory-selection mechanism, ensuring that our evaluation is both methodologically sound and consistent with the biological inspiration underlying SynapticRAG.

\paragraph{Equal Retrieval Count (ERC)}
However, fixed-$k$ retrieval becomes problematic as memory needs vary by context - from multiple memories for complex past events to few for simple queries. To address this, we propose ERC where retrieval count adapts dynamically: when our method retrieves $k$ memories for a query, other methods retrieve their top-$k$ ranked results, with accuracy calculated as the ratio of correct matches to total correct labels. Additionally, for queries requiring at least $m$ correct memories, we allow baselines to retrieve up to $m$ memories when the evaluated method retrieves fewer, denoted as Equal Retrieval Count with Minimum Guarantee, \textit{i.e.}, ERC-MG. The ERC-MG metric provides a stricter benchmark than ERC by ensuring baseline methods retrieve at least the annotated number of memories whenever SynapticRAG retrieves fewer. This setup rigorously evaluates SynapticRAG's robustness under conditions favorable to baseline methods. 

Notably, SynapticRAG achieves consistently high performance under the ERC-MG metric across all datasets, surpassing the strongest baseline by 7.76\%-14.66\%. Given that ERC-MG imposes more stringent retrieval requirements by allowing baseline methods the opportunity to retrieve all annotated memories, this result is particularly significant.

\subsection{Implementation Details}
Our implementation uses text-embedding-3-large \citep{openaiEmbeddings} for text vectorization and GPT-4o \citep{openai} for  generation. The memory management parameters include cosine similarity threshold cos$_{\text{th}}=0.262$ and temporal decay factor $\tau_{\text{init}}=43.07$. For the LIF model, we set membrane potential threshold $V_{\text{th}}=0.099$, stimulus threshold stim$_\text{th}=0.037$, and post-firing reset values $V_{\text{rest}}=2.903$, $I_{\text{rest}}=-7.128$. Complete hyperparameter settings are available in Table \ref{table:hyperparameters}.

\begin{table*}[ht]
\centering
\caption{Performance improvement of \ourmethod~and its modules ($\Delta$\%). }
\label{tab:ablation}
\begin{tabular}{l|cccc|c}
\toprule
Model Ablations & SMRCs$^{\text{EN}}$ & SMRCs$^{\text{JA}}$ & PerLTQA$^{\text{EN}}$ & PerLTQA$^{\text{CN}}$ & Avg. \\
\midrule
\textit{full} \ourmethod& \textbf{14.66} & \textbf{10.74} & \textbf{8.33} & \textbf{8.17} & \textbf{10.41} \\
\quad  \textit{w}/\textit{o} Temporal Integration & -11.82 & -8.55 & -9.68 & -9.54 & -9.90 \\
\quad  \textit{w}/\textit{o}  Stimulus Propagation & -8.31 & -2.41 & -3.60 & -3.36 & -4.35 \\
\quad  \textit{w}/\textit{o}  Dynamic Time Constant & -14.66 & -10.96 & -8.29 & -8.24 & -10.54\\
\bottomrule
\end{tabular}
\end{table*}

\begin{figure}[t]
\small
\centering
\includegraphics[width=0.98\columnwidth]
{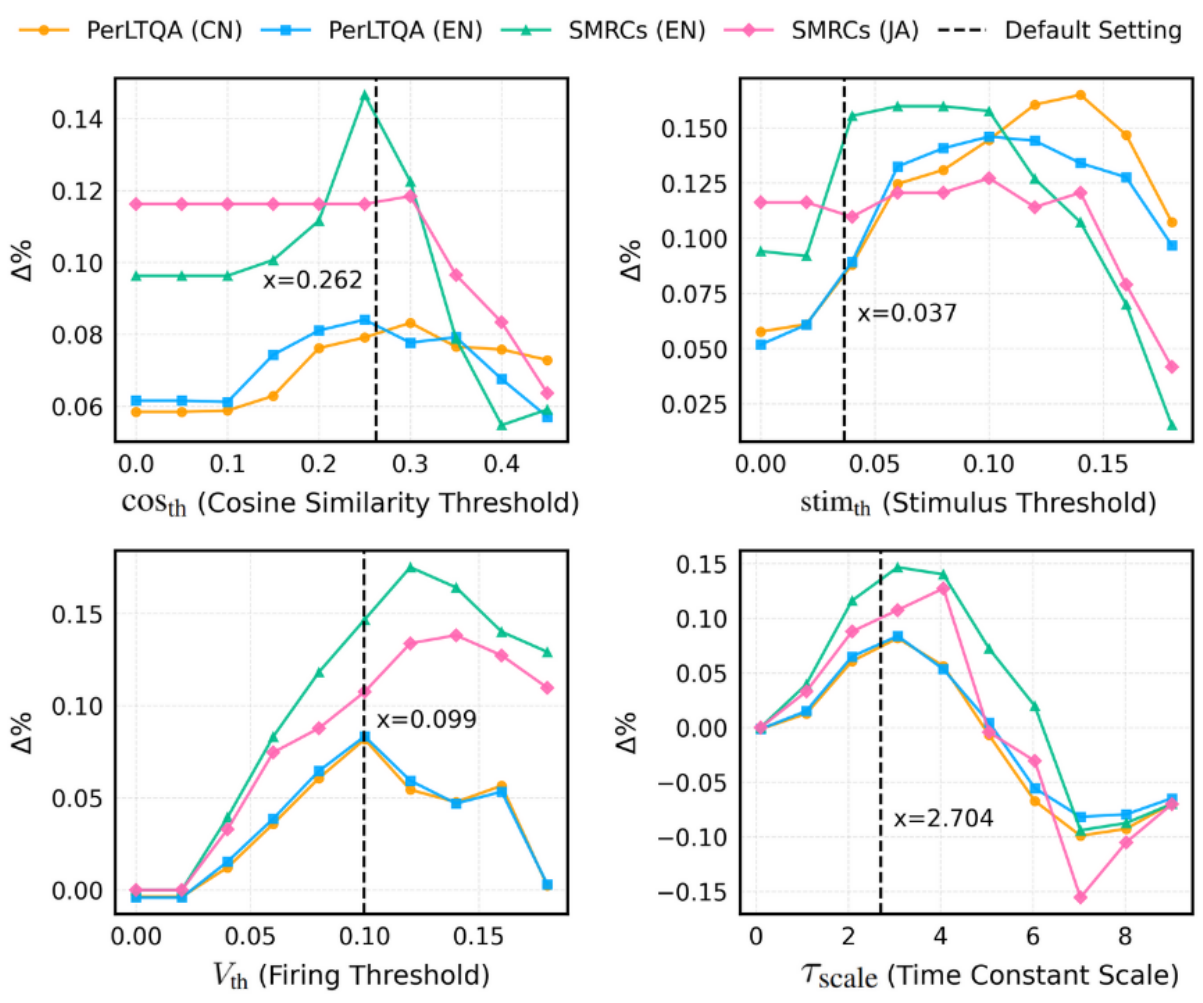}
\caption{Effect of key parameters on SynapticRAG's performance gain over RAG. $\Delta$\% indicates the accuracy improvement relative to RAG, and {Default Setting} denotes hyperparameter-optimal values.}
\label{fig:parameter}
\end{figure}

\section{Results}

\subsection{Main Results}
Experimental results shown in Figure \ref{fig:k_values} and Table \ref{tab:retrieval_performance} demonstrate that \ourmethod~achieves consistent performance improvements across four datasets: SMRCs (EN), SMRCs (JA), PerLTQA (EN) and PerLTQA (CN). Compared to the strongest baseline, our method shows an average improvement of 10.38\% in memory retrieval accuracy.

Specifically, from a cross-dataset perspective, \ourmethod~maintains stable advantages on both SMRCs with complex temporal dependencies and PerLTQA with natural dialogue flow. This consistency extends across languages (English, Japanese, and Chinese), validating our model's language-agnostic nature and generalization capability.

In terms of performance metrics, \ourmethod~successfully balances precision and recall, particularly excelling when $k\geqq6$, indicating superior handling of complex dialogues requiring multiple related memories. On SMRCs (EN), it achieves 86.21\% accuracy under ERC evaluation, surpassing the strongest baseline by 14.66\%. Similar improvements are observed across other datasets, with PerLTQA (CN) reaching 93.12\% accuracy.

The performance curves reveal that while precision decreases with increasing $k$ for all methods, \ourmethod~maintains a more gradual decline. Its recall curves consistently dominate other methods, demonstrating effective identification of relevant memories across temporal spans.

These results validate SynapticRAG's effectiveness in enhancing temporal memory retrieval, particularly in maintaining high recall while preserving precision. This balanced performance is especially valuable for applications requiring accurate memory retrieval across dialogue histories.

\begin{figure*}[ht]
\small
\centering
\includegraphics[width=0.98\textwidth]{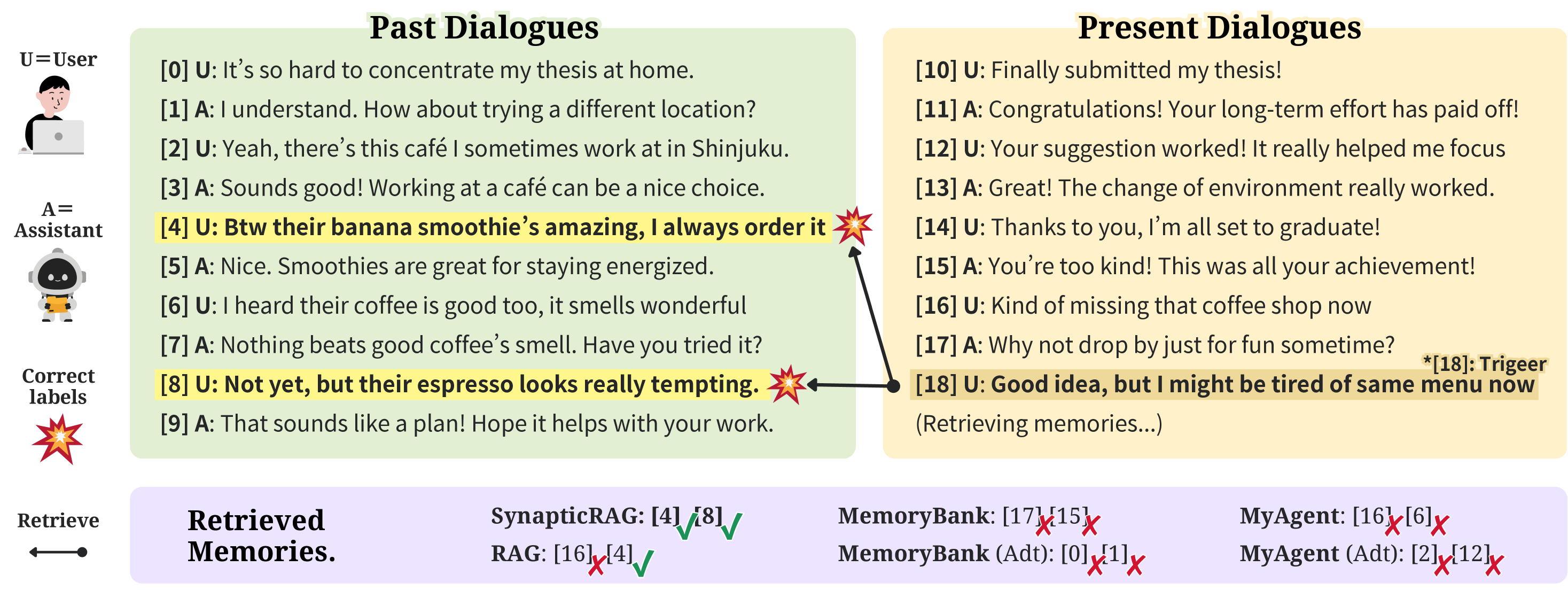}
\caption{A case study illustrating how different models retrieve memories.}
\label{fig:case}
\end{figure*}

\subsection{Ablation Study} \label{ablation}
Our method comprises three key components: a) Temporal Integration capturing temporal dependencies through DTW; b) Stimulus Propagation establishing synaptic-like memory connections through optimal propagation pathways; and c) Stimulus Activation controlling memory firing through dynamic LIF model, where decay factor $\tau$ balances recent and long-term memories. We evaluate each component's contribution through individual removal. Specifically, we set T$_{\text{score}}$=1 to remove temporal association (\textit{w}/\textit{o} Temporal Integration) for examining pure semantic similarity-based retrieval; disable stimulus propagation by setting stim$_{\text{th}}$=1 (\textit{w}/\textit{o} Stimulus Propagation) to observe performance when memories are independent and unaffected by each other's stimulation; and fix $\tau$ to its initial value $\tau_{\text{init}}$ (\textit{w}/\textit{o} Dynamic Time Constant) to examine the effect of prioritizing recent inputs without balancing long-term memory consideration. 

Results in Table \ref{tab:ablation} demonstrate that each module is crucial, with their removal leading to significant performance degradation across all datasets. Specifically, removing Temporal Integration causes an average performance drop of 9.90\%, indicating that semantic similarity alone struggles to capture complex temporal dependencies in dialogues; disabling Stimulus Propagation results in an average drop of 4.35\%, validating the necessity of memory interconnections; removing Dynamic Time Constant leads to substantial degradation with an average drop of 10.54\%, demonstrating the critical role of balancing recent inputs and long-term memories. These findings validate both our component design and biologically-inspired principles, particularly on SMRCs with complex temporal dependencies.

\subsection{Parameter Analysis}
Building upon our validation of individual components, we analyze SynapticRAG's parameter sensitivity through four fundamental parameters: cos$_{\text{th}}$ for consine similarity threshold , stim$_{\text{th}}$ for stimulus propagation threshold, $V_{\text{th}}$ for firing threshold, and $\tau_{\text{scale}}$ for scaling dynamic time constant. While Optuna optimization identified their optimal values, understanding their impacts remains crucial for both theoretical and practical purposes.

As shown in Figure \ref{fig:parameter}, our analysis reveals distinct parameter sensitivities. The filtering threshold cos$_{\text{th}}$ achieves optimal performance at $0.262$ within $[0.0, 0.4]$, with SMRCs showing higher sensitivity than PerLTQA due to their complex semantic relationships. The stimulus threshold stim$_{\text{th}}$ exhibits stable behavior around $0.037$ in $[0.0, 0.17]$, where higher values block essential connections while lower values cause over-propagation. The firing threshold $V_{\text{th}}$ shows moderate sensitivity at $0.099$ within $[0.0, 0.2]$, balancing between over-activation and insufficient utilization.

Most critically, $\tau_{\text{scale}}$ demonstrates dramatic performance variations, peaking at $2.704$ with 
a sharp decline beyond this value (negative $\Delta\%$). These findings suggest task-specific tuning strategies: precise adjustment for semantic-heavy tasks, stable propagation thresholds for general applications, and potentially adaptive mechanisms for temporal scaling. The identified robust regions can guide efficient parameter tuning in practical deployments.

\subsection{Case Study}
To demonstrate SynapticRAG's temporal memory capabilities, we present a case from SMRCs where a user discusses their thesis writing experience at a café (Figure \ref{fig:case}). When the user expresses potential menu fatigue at turn [18], \ourmethod~successfully retrieves the correct label memories about beverage preferences ([4] and [8]) from past dialogue, where they mentioned a banana smoothie and interest in espresso. This temporal association enables a contextually appropriate response that acknowledges both their past interest and current desire for change. RAG retrieves one correct memory ([4]) but misses the temporal connection ([16]), MemoryBank and MyAgent fail to retrieve either correct memory, and their adapted versions show similar temporal association failures. This example illustrates SynapticRAG's ability to form coherent memory chains across temporal gaps while maintaining contextual relevance, a key advantage of our synaptic-inspired propagation mechanism.

\section{Conclusion}
In this paper, we introduced SynapticRAG, a biologically-inspired memory retrieval model that enhances temporal memory management in dialogue systems. By integrating temporal association triggers, synaptic-like stimulus propagation, and dynamic memory activation mechanisms, our approach effectively addresses the limitations of simple similarity-based retrieval in handling temporally distributed conversations. Our experiments demonstrate while RAG excels in simpler straightforward retrieval tasks, SynapticRAG demonstrates superior performance in temporal-aware dialogue contexts where understanding dialogue history is critical, achieving consistent improvements across multiple languages and dialogue scenarios.

\section{Limitations}
\paragraph{Over-retrieval Risk.}
The stimulus propagation mechanism may trigger excessive memory node activations through synaptic connections, particularly in complex multi-topic dialogues. Such over-retrieval leads to two main issues: lengthy generation prompts as all retrieved memory texts are incorporated for response generation, and increased computational overhead due to linear growth of memory storage with dialogue length. To mitigate these issues, early stopping criteria and pruning strategies could be implemented to terminate unnecessary activation paths while preserving essential temporal connections, and importance sampling could be used to maintain only the most crucial historical information.

\paragraph{Computational Complexity.}
The original SynapticRAG architecture incurs computational complexity on the order of $O(n^2)$ in memory retrieval, where $n$ represents the total number of memory nodes. This quadratic complexity arises from the cosine-similarity filtering process, as each propagation step involves multiple parent nodes independently performing full scans through the entire memory database. To address this limitation, we have developed an optimized architecture that computes a single representative centroid vector based on all parent nodes, reducing the complexity from $O(n^2)$ to $O(n)$. See details in Appendix \ref{appendix:complexity_optimization}.

\paragraph{Parameter Sensitivity.}
The optimization of model parameters requires more extensive tuning compared to simpler models, and the variable number of recalled memories presents challenges when strict control is needed. Different language contexts may require separate parameter tuning. This could be mitigated through the development of automated parameter adaptation mechanisms that dynamically adjust based on input characteristics and language-specific features.

\paragraph{Model Interpretability.}
The complex stimulus propagation and firing mechanisms reduce interpretability compared to simpler similarity-based approaches. Tracing memory selection reasoning becomes challenging due to dynamic synaptic connections. However, this could potentially be solved by introducing attention-like mechanisms to quantify the contribution of each synaptic connection to the final memory selection, providing insights into the model's decision process.

\section*{Acknowledgments}
The work of Qinghua Zhao was partially supported by the Hefei College Talent Research Fund Project (No.
24RC20, 21-22RC13), and Natural Science Foundation of the AnHui Higher Education Institutions of China (No. 2022AH051779).

\bibliography{custom}

\begin{thebibliography}{47}
\providecommand{\natexlab}[1]{#1}

\bibitem[{Alonso et~al.(2024)Alonso, Figliolia, Ndirango, and Millidge}]{alonso2024conversationalagentscontexttime}
Nick Alonso, Tomás Figliolia, Anthony Ndirango, and Beren Millidge. 2024.
\newblock \href {https://arxiv.org/abs/2406.00057} {Toward conversational agents with context and time sensitive long-term memory}.
\newblock \emph{Preprint}, arXiv:2406.00057.

\bibitem[{Anthropic(2024)}]{claude3.5}
Anthropic. 2024.
\newblock Claude 3.5 sonnet.
\newblock \url{https://www.anthropic.com}.
\newblock Large language model, accessed February 3, 2025.

\bibitem[{Beltagy et~al.(2020)Beltagy, Peters, and Cohan}]{beltagy2020longformerlongdocumenttransformer}
Iz~Beltagy, Matthew~E. Peters, and Arman Cohan. 2020.
\newblock \href {https://arxiv.org/abs/2004.05150} {Longformer: The long-document transformer}.
\newblock \emph{Preprint}, arXiv:2004.05150.

\bibitem[{Borgeaud et~al.(2022)Borgeaud, Mensch, Hoffmann, Cai, Rutherford, Millican, van~den Driessche, Lespiau, Damoc, Clark, de~Las~Casas, Guy, Menick, Ring, Hennigan, Huang, Maggiore, Jones, Cassirer, Brock, Paganini, Irving, Vinyals, Osindero, Simonyan, Rae, Elsen, and Sifre}]{rag2}
Sebastian Borgeaud, Arthur Mensch, Jordan Hoffmann, Trevor Cai, Eliza Rutherford, Katie Millican, George van~den Driessche, Jean-Baptiste Lespiau, Bogdan Damoc, Aidan Clark, Diego de~Las~Casas, Aurelia Guy, Jacob Menick, Roman Ring, Tom Hennigan, Saffron Huang, Loren Maggiore, Chris Jones, Albin Cassirer, Andy Brock, Michela Paganini, Geoffrey Irving, Oriol Vinyals, Simon Osindero, Karen Simonyan, Jack~W. Rae, Erich Elsen, and Laurent Sifre. 2022.
\newblock \href {https://arxiv.org/abs/2112.04426} {Improving language models by retrieving from trillions of tokens}.
\newblock \emph{Preprint}, arXiv:2112.04426.

\bibitem[{Bracewell(2000)}]{bracewell2000fourier}
R.N. Bracewell. 2000.
\newblock \href {https://books.google.co.jp/books?id=ZNQQAQAAIAAJ} {\emph{The Fourier Transform and Its Applications}}.
\newblock Circuits and systems. McGraw Hill.

\bibitem[{Brown et~al.(2020)Brown, Mann, Ryder, Subbiah, Kaplan, Dhariwal, Neelakantan, Shyam, Sastry, Askell, Agarwal, Herbert-Voss, Krueger, Henighan, Child, Ramesh, Ziegler, Wu, Winter, Hesse, Chen, Sigler, Litwin, Gray, Chess, Clark, Berner, McCandlish, Radford, Sutskever, and Amodei}]{NEURIPS2020_1457c0d6}
Tom Brown, Benjamin Mann, Nick Ryder, Melanie Subbiah, Jared~D Kaplan, Prafulla Dhariwal, Arvind Neelakantan, Pranav Shyam, Girish Sastry, Amanda Askell, Sandhini Agarwal, Ariel Herbert-Voss, Gretchen Krueger, Tom Henighan, Rewon Child, Aditya Ramesh, Daniel Ziegler, Jeffrey Wu, Clemens Winter, Chris Hesse, Mark Chen, Eric Sigler, Mateusz Litwin, Scott Gray, Benjamin Chess, Jack Clark, Christopher Berner, Sam McCandlish, Alec Radford, Ilya Sutskever, and Dario Amodei. 2020.
\newblock \href {https://proceedings.neurips.cc/paper_files/paper/2020/file/1457c0d6bfcb4967418bfb8ac142f64a-Paper.pdf} {Language models are few-shot learners}.
\newblock In \emph{Advances in Neural Information Processing Systems}, volume~33, pages 1877--1901. Curran Associates, Inc.

\bibitem[{Brunel and Rossum(2007)}]{lif1}
Nicolas Brunel and Mark~C. Rossum. 2007.
\newblock Lapicque's 1907 paper: from frogs to integrate-and-fire.
\newblock \emph{Biol. Cybern.}, 97(5–6):337–339.

\bibitem[{Compte et~al.(2000)Compte, Brunel, Goldman-Rakic, and Wang}]{Compte2000}
Albert Compte, Nicolas Brunel, Patricia~S. Goldman-Rakic, and Xiao-Jing Wang. 2000.
\newblock \href {https://doi.org/10.1093/cercor/10.9.910} {Synaptic mechanisms and network dynamics underlying spatial working memory in a cortical network model}.
\newblock \emph{Cerebral Cortex}, 10(9):910--923.

\bibitem[{Croft et~al.(2009)Croft, Metzler, and Strohman}]{10.5555/1516224}
Bruce Croft, Donald Metzler, and Trevor Strohman. 2009.
\newblock \emph{Search Engines: Information Retrieval in Practice}, 1st edition.
\newblock Addison-Wesley Publishing Company, USA.

\bibitem[{Dai et~al.(2019)Dai, Yang, Yang, Carbonell, Le, and Salakhutdinov}]{Dai2019TransformerXLAL}
Zihang Dai, Zhilin Yang, Yiming Yang, Jaime~G. Carbonell, Quoc~V. Le, and Ruslan Salakhutdinov. 2019.
\newblock \href {https://api.semanticscholar.org/CorpusID:57759363} {Transformer-xl: Attentive language models beyond a fixed-length context}.
\newblock In \emph{Annual Meeting of the Association for Computational Linguistics}.

\bibitem[{Dayan and Abbott(2005)}]{lif2}
Peter Dayan and L.~F. Abbott. 2005.
\newblock \emph{Theoretical Neuroscience: Computational and Mathematical Modeling of Neural Systems}.
\newblock The MIT Press.

\bibitem[{Deng et~al.(2024)Deng, Li, Yu, and Ma}]{deng2024}
Xi~Deng, Shasha Li, Jie Yu, and Jun Ma. 2024.
\newblock \href {https://doi.org/10.1007/978-981-97-2421-5_16} {Hm-transformer: Hierarchical multi-modal transformer for long document image understanding}.
\newblock In \emph{Web and Big Data: 7th International Joint Conference, APWeb-WAIM 2023, Wuhan, China, October 6–8, 2023, Proceedings, Part IV}, page 232–245, Berlin, Heidelberg. Springer-Verlag.

\bibitem[{Du et~al.(2024)Du, Wang, Zhao, Liang, Wang, Zhong, Wang, and Wong}]{du-etal-2024-perltqa}
Yiming Du, Hongru Wang, Zhengyi Zhao, Bin Liang, Baojun Wang, Wanjun Zhong, Zezhong Wang, and Kam-Fai Wong. 2024.
\newblock \href {https://aclanthology.org/2024.sighan-1.18/} {{P}er{LTQA}: A personal long-term memory dataset for memory classification, retrieval, and fusion in question answering}.
\newblock In \emph{Proceedings of the 10th SIGHAN Workshop on Chinese Language Processing (SIGHAN-10)}, pages 152--164, Bangkok, Thailand. Association for Computational Linguistics.

\bibitem[{Dubrow and Davachi(2013)}]{tempmemorycite2}
Sarah Dubrow and Lila Davachi. 2013.
\newblock \href {https://api.semanticscholar.org/CorpusID:9932777} {The influence of context boundaries on memory for the sequential order of events.}
\newblock \emph{Journal of experimental psychology. General}, 142 4:1277--86.

\bibitem[{Fell and Axmacher(2011)}]{fell2011}
Juergen Fell and Nikolai Axmacher. 2011.
\newblock \href {https://doi.org/10.1038/nrn2979} {The role of phase synchronization in memory processes}.
\newblock \emph{Nature reviews. Neuroscience}, 12:105--18.

\bibitem[{Gao et~al.(2018)Gao, Galley, and Li}]{gao-etal-2018-neural-approaches}
Jianfeng Gao, Michel Galley, and Lihong Li. 2018.
\newblock \href {https://doi.org/10.18653/v1/P18-5002} {Neural approaches to conversational {AI}}.
\newblock In \emph{Proceedings of the 56th Annual Meeting of the Association for Computational Linguistics: Tutorial Abstracts}, pages 2--7, Melbourne, Australia. Association for Computational Linguistics.

\bibitem[{Gerstner and Kistler(2002)}]{Gerstner2002SPIKINGNM}
Wulfram Gerstner and Werner~M. Kistler. 2002.
\newblock \emph{Spiking Neuron Models: Single Neurons, Populations, Plasticity}.
\newblock Cambridge University Press.

\bibitem[{Graves et~al.(2016)Graves, Wayne, Reynolds, Harley, Danihelka, Grabska-Barwińska, Gómez, Grefenstette, Ramalho, Agapiou, Badia, Hermann, Zwols, Ostrovski, Cain, King, Summerfield, Blunsom, Kavukcuoglu, and Hassabis}]{graves2016hybrid}
Alex Graves, Greg Wayne, Malcolm Reynolds, Tim Harley, Ivo Danihelka, Agnieszka Grabska-Barwińska, Sergio Gómez, Edward Grefenstette, Tiago Ramalho, John Agapiou, Adrià Badia, Karl Hermann, Yori Zwols, Georg Ostrovski, Adam Cain, Helen King, Christopher Summerfield, Phil Blunsom, Koray Kavukcuoglu, and Demis Hassabis. 2016.
\newblock \href {https://doi.org/10.1038/nature20101} {Hybrid computing using a neural network with dynamic external memory}.
\newblock \emph{Nature}, 538.

\bibitem[{Gutiérrez et~al.(2025)Gutiérrez, Shu, Gu, Yasunaga, and Su}]{hipporag}
Bernal~Jiménez Gutiérrez, Yiheng Shu, Yu~Gu, Michihiro Yasunaga, and Yu~Su. 2025.
\newblock \href {https://arxiv.org/abs/2405.14831} {Hipporag: Neurobiologically inspired long-term memory for large language models}.
\newblock \emph{Preprint}, arXiv:2405.14831.

\bibitem[{Guu et~al.(2020)Guu, Lee, Tung, Pasupat, and Chang}]{rag3}
Kelvin Guu, Kenton Lee, Zora Tung, Panupong Pasupat, and Ming-Wei Chang. 2020.
\newblock \href {https://arxiv.org/abs/2002.08909} {Realm: Retrieval-augmented language model pre-training}.
\newblock \emph{Preprint}, arXiv:2002.08909.

\bibitem[{Hambarde and Proença(2023)}]{10184013}
Kailash~A. Hambarde and Hugo Proença. 2023.
\newblock \href {https://doi.org/10.1109/ACCESS.2023.3295776} {Information retrieval: Recent advances and beyond}.
\newblock \emph{IEEE Access}, 11:76581--76604.

\bibitem[{Hebb(1949)}]{hebb}
Donald~Olding Hebb. 1949.
\newblock \emph{The organization of behavior; a neuropsychological theory}.
\newblock Wiley.

\bibitem[{Hou et~al.(2024)Hou, Tamoto, and Miyashita}]{myagent}
Yuki Hou, Haruki Tamoto, and Homei Miyashita. 2024.
\newblock \href {https://doi.org/10.1145/3613905.3650839} {My agent understands me better: Integrating dynamic human-like memory recall and consolidation in llm-based agents}.
\newblock In \emph{Extended Abstracts of the CHI Conference on Human Factors in Computing Systems}, CHI EA '24, New York, NY, USA. Association for Computing Machinery.

\bibitem[{Howard et~al.(2024)Howard, Esfahani, Le, and Sederberg}]{Howard2024}
Marc~W. Howard, Zahra~G. Esfahani, Bao Le, and Per~B. Sederberg. 2024.
\newblock \href {https://arxiv.org/abs/2302.10163} {Learning temporal relationships between symbols with laplace neural manifolds}.
\newblock \emph{Preprint}, arXiv:2302.10163.

\bibitem[{Howard and Kahana(2002)}]{tempmemorycite1}
Marc~W. Howard and Michael~J. Kahana. 2002.
\newblock \href {https://doi.org/10.1006/jmps.2001.1388} {A distributed representation of temporal context}.
\newblock \emph{Journal of Mathematical Psychology}, 46(3):269--299.

\bibitem[{Krippendorff(2004)}]{krippendorff04}
Klaus Krippendorff. 2004.
\newblock \emph{Content Analysis: An Introduction to Its Methodology (second edition)}.
\newblock Sage Publications.

\bibitem[{Krippendorff(2011)}]{krippendorff2011computing}
Klaus Krippendorff. 2011.
\newblock \href {https://api.semanticscholar.org/CorpusID:59901023} {Computing krippendorff's alpha-reliability}.

\bibitem[{Lee et~al.(2024)Lee, Park, Hong, Kim, Chang, and Choi}]{lee-etal-2024-improving-conversational}
Janghwan Lee, Seongmin Park, Sukjin Hong, Minsoo Kim, Du-Seong Chang, and Jungwook Choi. 2024.
\newblock \href {https://doi.org/10.18653/v1/2024.acl-long.612} {Improving conversational abilities of quantized large language models via direct preference alignment}.
\newblock In \emph{Proceedings of the 62nd Annual Meeting of the Association for Computational Linguistics (Volume 1: Long Papers)}, pages 11346--11364, Bangkok, Thailand. Association for Computational Linguistics.

\bibitem[{Lewis et~al.(2020)Lewis, Perez, Piktus, Petroni, Karpukhin, Goyal, K\"{u}ttler, Lewis, Yih, Rockt\"{a}schel, Riedel, and Kiela}]{rag}
Patrick Lewis, Ethan Perez, Aleksandra Piktus, Fabio Petroni, Vladimir Karpukhin, Naman Goyal, Heinrich K\"{u}ttler, Mike Lewis, Wen-tau Yih, Tim Rockt\"{a}schel, Sebastian Riedel, and Douwe Kiela. 2020.
\newblock Retrieval-augmented generation for knowledge-intensive nlp tasks.
\newblock In \emph{Proceedings of the 34th International Conference on Neural Information Processing Systems}, NIPS '20. Curran Associates Inc.

\bibitem[{Manning et~al.(2008)Manning, Raghavan, and Schütze}]{Manning_Raghavan_Schütze_2008}
Christopher~D. Manning, Prabhakar Raghavan, and Hinrich Schütze. 2008.
\newblock \emph{Introduction to Information Retrieval}.
\newblock Cambridge University Press.

\bibitem[{McDaniel and Einstein(2000)}]{Mcdanie}
Mark~A McDaniel and Gilles~O Einstein. 2000.
\newblock Strategic and automatic processes in prospective memory retrieval: a multiprocess framework.
\newblock \emph{Applied Cognitive Psychology}, 14.

\bibitem[{Neves et~al.(2008)Neves, Cooke, and Bliss}]{neves2008synaptic}
Guilherme Neves, Sam Cooke, and Tim Bliss. 2008.
\newblock \href {https://doi.org/10.1038/nrn2303} {Synaptic plasticity, memory and the hippocampus: a neural network approach to causality}.
\newblock \emph{Nature Reviews Neuroscience}, 9:65--75.

\bibitem[{Norman and O’Reilly(2003)}]{Norman2003}
Kenneth Norman and Randall O’Reilly. 2003.
\newblock \href {https://doi.org/10.1037/0033-295X.110.4.611} {Modeling hippocampal and neocortical contributions to recognition memory: A complementary learning systems approach}.
\newblock \emph{Psychological review}, 110:611--46.

\bibitem[{Olabiyi et~al.(2019)Olabiyi, Salimov, Khazane, and Mueller}]{olabiyi-etal-2019-multi}
Oluwatobi Olabiyi, Alan~O Salimov, Anish Khazane, and Erik Mueller. 2019.
\newblock \href {https://doi.org/10.18653/v1/W19-4114} {Multi-turn dialogue response generation in an adversarial learning framework}.
\newblock In \emph{Proceedings of the First Workshop on NLP for Conversational AI}, pages 121--132, Florence, Italy. Association for Computational Linguistics.

\bibitem[{OpenAI(2024{\natexlab{a}})}]{openaiEmbeddings}
OpenAI. 2024{\natexlab{a}}.
\newblock Embeddings - openai api.
\newblock \url{https://platform.openai.com/docs/guides/embeddings}.
\newblock (Accessed on 05/21/2024).

\bibitem[{OpenAI(2024{\natexlab{b}})}]{openai}
OpenAI. 2024{\natexlab{b}}.
\newblock \href {https://arxiv.org/abs/2410.21276} {Gpt-4o system card}.
\newblock \emph{Preprint}, arXiv:2410.21276.

\bibitem[{O’Reilly and McClelland(1994)}]{OReilly1994HippocampalCE}
Randall~C. O’Reilly and James~L. McClelland. 1994.
\newblock \href {https://api.semanticscholar.org/CorpusID:9049086} {Hippocampal conjunctive encoding, storage, and recall: Avoiding a trade‐off}.
\newblock \emph{Hippocampus}, 4.

\bibitem[{Sakoe and Chiba(1978)}]{dtwJP}
H~Sakoe and S~Chiba. 1978.
\newblock \href {https://doi.org/10.1109/tassp.1978.1163055} {Dynamic programming algorithm optimization for spoken word recognition}.
\newblock \emph{IEEE Transactions on Acoustics, Speech, and Signal Processing}, 26(1):43--49.

\bibitem[{Sukhbaatar et~al.(2015)Sukhbaatar, Szlam, Weston, and Fergus}]{endtoend}
Sainbayar Sukhbaatar, Arthur Szlam, Jason Weston, and Rob Fergus. 2015.
\newblock End-to-end memory networks.
\newblock In \emph{Proceedings of the 29th International Conference on Neural Information Processing Systems - Volume 2}, NIPS'15, page 2440–2448, Cambridge, MA, USA. MIT Press.

\bibitem[{Trivedi et~al.(2023)Trivedi, Balasubramanian, Khot, and Sabharwal}]{trivedi-etal-2023-interleaving}
Harsh Trivedi, Niranjan Balasubramanian, Tushar Khot, and Ashish Sabharwal. 2023.
\newblock \href {https://doi.org/10.18653/v1/2023.acl-long.557} {Interleaving retrieval with chain-of-thought reasoning for knowledge-intensive multi-step questions}.
\newblock In \emph{Proceedings of the 61st Annual Meeting of the Association for Computational Linguistics (Volume 1: Long Papers)}, pages 10014--10037, Toronto, Canada. Association for Computational Linguistics.

\bibitem[{Tulving(1985)}]{Tul1985}
Endel Tulving. 1985.
\newblock \emph{Elements of episodic memory}.
\newblock Number~2 in Oxford psychology series. Clarendon Press, Oxford University Press.

\bibitem[{Wang(2001)}]{Wang2001}
Xiao-Jing Wang. 2001.
\newblock \href {https://api.semanticscholar.org/CorpusID:9843447} {Synaptic reverberation underlying mnemonic persistent activity}.
\newblock \emph{Trends in Neurosciences}, 24:455--463.

\bibitem[{Wang et~al.(2024)Wang, Liu, Chen, O'Brien, Wu, and McAuley}]{liu2024self}
Yu~Wang, Xinshuang Liu, Xiusi Chen, Sean O'Brien, Junda Wu, and Julian~J. McAuley. 2024.
\newblock \href {https://api.semanticscholar.org/CorpusID:273023170} {Self-updatable large language models with parameter integration}.
\newblock \emph{ArXiv}, abs/2410.00487.

\bibitem[{Wu(2019)}]{wu2019learning}
Chien-Sheng Wu. 2019.
\newblock \href {https://api.semanticscholar.org/CorpusID:159040674} {Learning to memorize in neural task-oriented dialogue systems}.
\newblock \emph{ArXiv}, abs/1905.07687.

\bibitem[{Zaheer et~al.(2020)Zaheer, Guruganesh, Dubey, Ainslie, Alberti, Ontanon, Pham, Ravula, Wang, Yang, and Ahmed}]{NEURIPS2020_c8512d14}
Manzil Zaheer, Guru Guruganesh, Kumar~Avinava Dubey, Joshua Ainslie, Chris Alberti, Santiago Ontanon, Philip Pham, Anirudh Ravula, Qifan Wang, Li~Yang, and Amr Ahmed. 2020.
\newblock \href {https://proceedings.neurips.cc/paper_files/paper/2020/file/c8512d142a2d849725f31a9a7a361ab9-Paper.pdf} {Big bird: Transformers for longer sequences}.
\newblock In \emph{Advances in Neural Information Processing Systems}, volume~33, pages 17283--17297. Curran Associates, Inc.

\bibitem[{Zhang et~al.(2019)Zhang, Huang, Zhao, Ji, Chen, and Zhu}]{zhang2018memory}
Zheng Zhang, Minlie Huang, Zhongzhou Zhao, Feng Ji, Haiqing Chen, and Xiaoyan Zhu. 2019.
\newblock \href {https://doi.org/10.1145/3317612} {Memory-augmented dialogue management for task-oriented dialogue systems}.
\newblock \emph{ACM Trans. Inf. Syst.}, 37(3).

\bibitem[{Zhong et~al.(2023)Zhong, Guo, Gao, Ye, and Wang}]{memorybank}
Wanjun Zhong, Lianghong Guo, Qiqi Gao, He~Ye, and Yanlin Wang. 2023.
\newblock \href {https://arxiv.org/abs/2305.10250} {Memorybank: Enhancing large language models with long-term memory}.
\newblock \emph{Preprint}, arXiv:2305.10250.

\end{thebibliography}

\appendix

\section{SMRCs Dataset Generation Guidelines}
\label{guidelines}
We used the following prompt to generate the SMRCs dataset:

\begin{tcolorbox}[
    colback=gray!5,
    colframe=gray!40,
    boxrule=0.5pt,
    arc=0pt,
    left=10pt,
    right=10pt,
    top=10pt,
    bottom=10pt
]
\small
\textbf{Please generate a dialogue dataset following these guidelines:}

\textbf{1. Temporal Structure:}

- Create conversations at two time points: past (1-6 months ago) and present

- Each conversation set should contain approximately 10 turns of dialogue

\textbf{2. Conversation Design:}

- Topics should cover a broad range including daily life, work, school, and personal consultations

- Focus on personalized conversations rather than generic exchanges

- Keep each turn concise, limiting to 10-20 phrases per interaction

\textbf{3. Memory Retrieval Design:}

- The assistant should naturally recall past conversations based on the user's current statements

- Assistant's responses should reference specific recalled utterances from either the user or assistant

- Responses must maintain logical coherence with the recalled memories

\textbf{4. Memory Trigger Requirements:}

- Each trigger should be a single user utterance from the present conversation

- Recalled memories should be complete utterances from past conversations (either user or assistant)

- Multiple trigger-memory pairs within a single conversation are preferred

\textbf{5. Quality Criteria:}

- Ensure natural conversation flow with appropriate memory recalls

- Prioritize conceptual relevance over simple keyword matching

- Maintain clear and appropriate relationships between triggers and recalled memories

\textbf{6. Documentation Requirements:}\\
For each conversation set, explicitly document:

- Trigger: The specific user utterance in the present conversation that prompts recall

- Recalled Memories: The complete related utterances from past conversations

- All memories should be documented without abbreviation

\textbf{7. Output Format:}

- Provide results in JSON format only

\textbf{Note:} The focus should be on creating natural dialogue flows where memory recalls are contextually appropriate and based on potential conceptual relationships rather than simple pattern matching.
\end{tcolorbox}

\section{Hyperparameter Optimization} 
\label{hyperparameterOpt}
We optimized hyperparameters using Optuna, ensuring the evaluation metric was independent of comparative models. Optuna iteratively samples parameters, evaluates performance, and refines sampling to efficiently explore the parameter space and identify optimal configurations, eliminating the need for exhaustive manual tuning. When multiple trials achieved the same best score, we selected the parameter set with the highest ERC score to ensure robust performance. The number of hyperparameters varied by model: MemoryBank (0), MemoryBank(Adapted) (5), MyAgent (1), MyAgent(Adapted) (4), and \ourmethod~(9). 

\subsection{Optimization for Adapted Models}  
For adapted models, where the number of retrieved memories cannot be predetermined, we optimized to maximize the scores of correct label memories. Each memory score $s_i$ was normalized to [0, 1] using the total score sum $Sum_n = \sum_{i=1}^{M_n} s_i$, with $M_n$ being the number of memories in the $n$-th task. The normalized scores of correct label memories, indexed by $y_n$, were summed as $Score_n = \sum_{j \in y_n} s_j / Sum_n$. The final score across all $N$ tasks, $Score = \sum_{n=1}^N Score_n$, was maximized by optimizing hyperparameters.

\subsection{Optimization for SynapticRAG}
For SynapticRAG, which uses the score $v$ for threshold judgment, optimization was based on conditions related to fired memories. Two conditions were imposed. The first penalized insufficient correct label memories, defined as $A_n = \frac{|y_n| - |f_n \cap y_n|}{|y_n|}$ if $|y_n| > |f_n \cap y_n|$, and 0 otherwise, where $f_n$ is the set of fired memory indices and $y_n$ is the set of correct label indices. The second penalized excessive memory retrieval, expressed as $B_n = \frac{|f_n|}{|y_n|}$. The final score was calculated as $Score = \sum_{n=1}^N -(A_n + 0.15 * B_n)$. The factor 0.15 balanced the firing of correct label memories and excessive retrieval, determined through extensive experimentation. A larger value prioritizes precision, while a smaller value emphasizes recall. The negative sum avoided division by zero issues. Hyperparameters were optimized to maximize this score.

\section{Hyperparameter Settings}
\label{appendixB}
All hyperparameters are listed in Table \ref{table:hyperparameters}.

\subsection{SynapticRAG}
In our model, in addition to cos$_{\text{th}}$, V$_{\text{th}}$, stim$_{\text{th}}$, \( \tau_{\text{init}} \), \( V_{\text{rest}} \), \( I_{\text{rest}} \) mentioned in the text, we include scaling constants \( \tau_{\text{scale}} \) for adjusting $\tau$, \( t_{\text{scale}} \) for adjusting $t$ and \( P_{\text{scale}} \) for adjusting $P_{\text{score}}$ as targets for hyperparameter optimization.

Using these scaling constants, $\tau$, $t$, and $P_{\text{score}}$ are rewritten as follows:

\[
\tau \rightarrow \tau_{\text{scale}} \tau
\]
\[
t \rightarrow t_{\text{scale}} t
\]
\[
P_{\text{score}} \rightarrow P_{\text{scale}} P_{\text{score}}
\]

In summary, the following nine parameters are subject to hyperparameter optimization in our model:
cos$_{\text{th}}$, V$_{\text{th}}$, stim$_{\text{th}}$, \( \tau_{\text{init}} \), \( V_{\text{rest}} \), \( I_{\text{rest}} \), \( \tau_{\text{scale}} \), \( t_{\text{scale}} \), \( P_{\text{scale}} \).

\subsection{MemoryBank (Adapted)}
The original MemoryBank model calculates memory scores as follows:

First, for each memory:
\[
\text{Score}^{(i)} = \exp\left(\frac{\Delta t^{(i)}}{5s^{(i)}}\right)
\]

where \( \Delta t^{(i)} \) is the time difference between when the memory occurred and the current time, and \( s^{(i)} \) is the memory strength, with an initial value of 1.

Subsequently, the Score is cut off by a randomly set threshold \( \theta \), and memories below this are considered forgotten. We consider the set of memories that exceed the threshold:

\[
\text{Retain} = \{i \mid \text{Score}^{(i)} \geq \theta\}
\]

Based on the cosine similarity with the input vector, the top 6 memories are selected from this set:
\[
\text{Top}_6 = \text{argsort}_{i \in \text{Retain}}(\cos_{\text{sim}}(i))[:6]
\]

The strength $s$ of the selected memories is updated by adding 1:

\[
s^{(i)}_0 = 1, \quad s^{(i)}_{n} = s^{(i)}_{n-1} + 1 \quad \text{for} \quad i \in \text{Top}_6
\]

The memory strength is updated in this way at each turn.
In the Adapted model, we introduced five parameters to this model:
\( \text{Top}_k \), \( \Theta \), \( t_{\text{scale}} \), \( s_{\text{scale}} \), \( s_{\text{init}} \).

The equations are as follows:
\[
\text{Score}^{(i)} = \exp\left(\frac{t_{\text{scale}}\Delta t^{(i)}}{s_{\text{scale}} s_{i}}\right)
\]

\[
\text{Retain} = \{i \mid \text{Score}^{(i)} \geq \Theta\}
\]

\[
\text{Top}_k = \text{argsort}_{i \in \text{Retain}}(\cos_{\text{sim}}(i))[:k]
\]

\[
s^{(i)}_0 = s_{\text{init}}, \quad s^{(i)}_{n} = s^{(i)}_{n-1} + 1 \quad \text{for} \quad i \in \text{Top}_k
\]

The optimization was performed based on these set parameters.

\begin{table}[h]
\centering
\resizebox{!}{0.6\columnwidth}{
\begin{tabular}{l|l|l}
\toprule
\textbf{Model} & \textbf{Parameter} & \textbf{Value} \\
\midrule
{MemoryBank (Adt)} & Top$_k$ & 7 \\
 & $\Theta$ & 0.12 \\
 & $t_{\text{scale}}$ & 13.82 \\
 & $s_{\text{scale}}$ & 9.06 \\
 & $s_{\text{init}}$ & 0.05 \\
\hline
MyAgent & cos$_{\text{th}}$ & 0.30 \\
\hline
{MyAgent (Adt)} & cos$_{\text{th}}$ & 0.49 \\
 & $r_{\text{scale}}$ & 10.07 \\
 & $t_{\text{scale}}$ & -1.72 \\
 & $g_{\text{scale}}$ & -2.05 \\
\hline
{SynapticRAG} & cos$_{\text{th}}$ & 0.26 \\
 & $V_{\text{th}}$ & 0.01 \\
 & stim$_{\text{th}}$ & 0.04 \\
 & $\tau_{\text{init}}$ & 43.07 \\
 & $\tau_{\text{scale}}$ & 2.70 \\
 & $t_{\text{scale}}$ & 7.93 \\
 & $P_{\text{scale}}$ & 0.46 \\
 & $V_{\text{rest}}$ & 2.90 \\
 & $I_{\text{rest}}$ & -7.13 \\
\bottomrule
\end{tabular}
}
\caption{Hyperparameter Settings}
\label{table:hyperparameters}
\end{table}

\subsection{MyAgent (Adapted)}
The original MyAgent model is defined as follows:
First, the recall probability $p$ is evaluated for memories exceeding a cosine similarity threshold with the input vector:
\[
\text{Retain} = \{i \mid \cos_{\text{sim}}(i) \geq \cos_{\text{th}}\}
\]

$p$ is calculated as:
\[
\begin{array}{c}
p^{(i)}(t) = \dfrac{1 - \exp\Big(-\dfrac{r^{(i)}e^{-\Delta t^{(i)}}}{g^{(i)}}\Big)}{1 - e^{-1}} \\
\text{for } i \in \text{Retain}
\end{array}
\]
where \( \Delta t \) is the time difference between memory occurrence and current time, and \( g \) is memory strength, initially set to 1.
The memory with the highest p is retrieved, and its strength g is updated:
\[
j = \text{argmax}_{i \in \text{Retain}} (p^{(i)}(t))
\]
\[
g^{(j)}_0 = 1, \quad g^{(j)}_n = g^{(j)}_{n - 1} + \frac{1 - e^{-\Delta t^{(j)}}}{1 + e^{-\Delta t^{(j)}}} 
\]

The memory strength is thus updated in each turn.
The Adapted model introduces four parameters: cos$_{\text{th}}$, \( r_{\text{scale}} \), \( t_{\text{scale}} \), \( g_{\text{scale}} \).
cos$_{\text{th}}$ is the cosine similarity threshold from the original model.

The equations for other parameters are as follows:
\[
\text{Retain} = \{i \mid \cos_{\text{sim}}(i) \geq \cos_{\text{th}}\}
\]
\[
p^{(i)}(t) = \frac{1 - \exp(-r_{\text{scale}}r^{(i)} e^{-t_{\text{scale}}\Delta t^{(i)}} / {g_{\text{scale}}g^{(i)}})}{1 - e^{-1}}
\]
\[
j = \text{argmax}_{i \in \text{Retain}} (p^{(i)}(t))
\]
\[
g^{(j)}_0 = 1, \quad g^{(j)}_n = g^{(j)}_{n - 1} + \frac{1 - e^{-t_{\text{scale}}\Delta t^{(j)}}}{1 + e^{-t_{\text{scale}}\Delta t^{(j)}}} 
\]
Optimization was performed based on these parameters.

\section{Computational Complexity Optimization}
\label{appendix:complexity_optimization}
As we noted in section Limitations, the original SynapticRAG architecture indeed incurs computational complexity on the order of $O(n^2)$ in the worst case. The main reason for this quadratic complexity arises from the cosine-similarity filtering process used during memory retrieval. Specifically, in the original method, each propagation step involves selecting parent nodes (whose number we denote as $(1/a)n$), proportional to the total number of memory nodes $n$). Each of these parent nodes independently performs a full scan through the entire memory database, resulting in a total computational complexity of approximately $(1/a)n \times n = (1/a)n^2$.

To directly address this issue, we have developed and experimentally validated an optimized alternative architecture. Rather than allowing each parent node to individually scan the entire database, our optimized method computes a \textbf{single representative centroid vector} based on \textbf{all parent nodes}. This centroid vector is then used to perform only one cosine-similarity scan through the entire dataset.

The computational complexity of this optimized approach can be explained as follows: calculating the centroid vector involves processing $(1/a)n$ parent nodes, thus incurring a computational cost proportional to $(1/a)n$. Subsequently, performing a single cosine-similarity scan using this centroid vector across all memory nodes incurs an additional cost of $n$. Hence, the total computational complexity of the optimized architecture becomes $(1/a)n + n = (1 + 1/a)n$, effectively \textbf{reducing the complexity from $O(n^2)$ to $O(n)$}.


\begin{table*}[ht]
\centering
\caption{Retrieval Accuracy Comparison (ERC \& ERC-MG scores)}
\small
\begin{tabular}{lccccccccc}
\toprule
\multirow{2}{*}{Model} & \multicolumn{2}{c}{SMRCs(EN)} & \multicolumn{2}{c}{SMRCs(JA)} & \multicolumn{2}{c}{PerLTQA(EN)} & \multicolumn{2}{c}{PerLTQA(CN)} & \multirow{2}{*}{Avg.} \\
\cmidrule(lr){2-3} \cmidrule(lr){4-5} \cmidrule(lr){6-7} \cmidrule(lr){8-9}
& ERC & ERC-MG & ERC & ERC-MG & ERC & ERC-MG & ERC & ERC-MG & \\
\midrule
RAG & 71.55 & 71.55 & 81.58 & 81.58 & 82.14 & 82.51 & 84.95 & 85.36 & 80.15 \\
MemoryBank (Adt) & 62.36 & 62.58 & 72.15 & 72.15 & 67.24 & 67.54 & 71.31 & 71.61 & 68.37 \\
MyAgent (Adt) & 63.24 & 63.46 & 77.19 & 77.19 & 66.49 & 66.64 & 69.54 & 69.76 & 69.19 \\
\midrule
\textbf{SynapticRAG} & & & & & & & & & \\
\textbf{(Original)} & \textbf{86.21} & \textbf{86.21} & \textbf{92.32} & \textbf{92.32} & \textbf{90.47} & \textbf{90.47} & \textbf{93.12} & \textbf{93.12} & \textbf{90.53} \\
\bottomrule
\end{tabular}
\end{table*}


\begin{table*}[ht]
\centering
\caption{Optimized Architecture Results}
\small
\begin{tabular}{lccccccccc}
\toprule
\multirow{2}{*}{Model} & \multicolumn{2}{c}{SMRCs(EN)} & \multicolumn{2}{c}{SMRCs(JA)} & \multicolumn{2}{c}{PerLTQA(EN)} & \multicolumn{2}{c}{PerLTQA(CN)} & \multirow{2}{*}{Avg.} \\
\cmidrule(lr){2-3} \cmidrule(lr){4-5} \cmidrule(lr){6-7} \cmidrule(lr){8-9}
& ERC & ERC-MG & ERC & ERC-MG & ERC & ERC-MG & ERC & ERC-MG & \\
\midrule
RAG & 66.30 & 66.30 & 69.52 & 69.52 & 73.77 & 74.15 & 74.68 & 75.08 & 71.07 \\
MemoryBank (Adt) & 57.77 & 58.21 & 62.50 & 62.50 & 64.13 & 64.43 & 66.91 & 67.21 & 62.96 \\
MyAgent (Adt) & 55.14 & 55.14 & 58.11 & 58.11 & 53.62 & 53.77 & 54.05 & 54.27 & 55.28 \\
\midrule
\textbf{SynapticRAG} & & & & & & & & & \\
\textbf{(Optimized)} & \textbf{79.43} & \textbf{79.43} & \textbf{83.11} & \textbf{83.11} & \textbf{85.74} & \textbf{85.74} & \textbf{86.95} & \textbf{86.95} & \textbf{83.81} \\
\bottomrule
\end{tabular}
\end{table*}

\begin{table}[ht]
\centering
\caption{Execution Time Comparison}
\small
\begin{tabular}{lcc}
\toprule
Model & Original (s) & Optimized (s) \\
\midrule
RAG & 0.00147 & 0.00137 \\
MemoryBank (Adt) & 0.08099 & 0.07916 \\
MyAgent (Adt) & 0.04962 & 0.04948 \\
\textbf{SynapticRAG} & \textbf{0.14534} & \textbf{0.10801} \\
\bottomrule
\end{tabular}
\end{table}

We further analyzed the execution time differences on the SMRCs(EN) dataset. The execution time reported here represents average CPU processing time per retrieval task. Slight variations observed in baseline methods (except SynapticRAG) are mainly attributable to minor differences in the number of retrieved memories and computational overhead, but these variations remain within a negligible margin of error. Moreover, the optimized architecture achieves approximately a 1.3 $\times$ foldcrease in execution speed compared to the original SynapticRAG. Given the difference in computational complexity (from $O(n^2)$ to $O(n)$), we expect this performance improvement to become significantly more pronounced as the dataset size increases.

\end{document}